\documentclass[11pt,a4paper]{article}
\usepackage{times,latexsym}
\usepackage{url}
\usepackage[T1]{fontenc}

\PassOptionsToPackage{dvipsnames}{xcolor}

\usepackage{xspace,mfirstuc,tabulary}

\newif\iftaclinstructions
\taclinstructionsfalse 
\iftaclinstructions
\renewcommand{\confidential}{}
\renewcommand{\anonsubtext}{(No author info supplied here, for consistency with
TACL-submission anonymization requirements)}
\newcommand{\instr}
\fi

\usepackage{amsmath, amssymb, amsfonts}
\usepackage{multirow}
\usepackage{times}
\usepackage{latexsym}
\usepackage{graphicx}
\usepackage{makecell}
\usepackage{tabularx}
\usepackage{adjustbox}
\usepackage{sidecap}
\usepackage{soul}
\usepackage{graphicx}
\usepackage{tikz}
\usetikzlibrary{shadows}
\usetikzlibrary{shapes.geometric, arrows}

\RequirePackage[most,breakable]{tcolorbox}

\RequirePackage[color=yellow!20,linecolor=red,textsize=scriptsize]{todonotes}
\RequirePackage[inline,shortlabels]{enumitem}
\setlist{nosep}
\usepackage[preprint]{acl}

\usepackage{times}
\usepackage{latexsym}

\usepackage[T1]{fontenc}

\usepackage[utf8]{inputenc}

\def\shortname{\texttt{LM\textsuperscript{2}}}

\usepackage{microtype}

\usepackage{inconsolata}

\title{\shortname{}: A Simple Society of Language Models Solves Complex Reasoning}

\newcommand*\samethanks[1][\value{footnote}]{\footnotemark[#1]}
\newcommand*\circled[1]{\tikz[baseline=(char.base)]{
            \node[shape=circle,draw,inner sep=1pt] (char) {#1};}}
\author{Gurusha Juneja\thanks{Equal contribution} \\
  Microsoft Research, India \\
  \texttt{gurushajuneja@gmail.com} \\\And
  Subhabrata Dutta\samethanks \\
  IIT Delhi, India\\
  \texttt{subha0009@gmail.com} \\\And
  Tanmoy Chakraborty \\
  IIT Delhi, India \\
  \texttt{tanchak@ee.iitd.ac.in}}

\begin{document}
\maketitle
\begin{abstract}
Despite demonstrating emergent reasoning abilities, Large Language Models (LLMS) often lose track of complex, multi-step reasoning. Existing studies show that providing guidance via decomposing the original question into multiple subproblems elicits more robustness in LLM reasoning -- a decomposer generates the subproblems, and a solver solves each of these subproblems. However, these techniques fail to accommodate coordination between the decomposer and the solver modules (either in a single model or different specialized ones) -- the decomposer does not keep track of the ability of the solver to follow the decomposed reasoning. In this paper, we propose \shortname{} to address these challenges. \shortname{} modularizes the decomposition, solution, and verification into three different language models. The decomposer module identifies the key concepts necessary to solve the problem and generates step-by-step subquestions according to the reasoning requirement. The solver model generates the solution to the subproblems that are then checked by the verifier module; depending upon the feedback from the verifier, the reasoning context is constructed using the subproblems and the solutions. These models are trained to coordinate using policy learning. Exhaustive experimentation suggests the superiority of \shortname{} over existing methods on in- and out-domain reasoning problems, outperforming the best baselines by 8.1\% on MATH, 7.71\% on JEEBench, and 9.7\% on MedQA problems (code available at \url{https://github.com/LCS2-IIITD/Language_Model_Multiplex}).
\end{abstract}

\section{Introduction}
\label{sec:intro}

Recent trends in solving complex reasoning tasks using Large Language Models (LLMs) typically follow two different dominant approaches: (i) well-curated prompting techniques~\citep{PHP, yao2024tree} on LLMs of exorbitant size like GPT-4~\citep{openai2023gpt4}, or (ii) finetuning a relatively smaller LLM using domain-focused data~\citep{shao2024deepseekmath, toshniwal2024openmathinstruct, dutta2024frugal}. Methods from the former category heavily rely on the proprietary LLM being used and are prone to fail absolutely when employed with less powerful models. The latter category, though cost-effective compared to humongous LLMs, often loses in generalizability due to a narrow training domain. 

{\bf The chronicle of decomposed reasoning.} A number of recent literature has pointed out that LLMs tend to perform better on complex reasoning 
tasks when the problem is decomposed into step-by-step subproblems~\citep{least-to-most, DSP, juneja-etal-2023-small}. Earlier techniques demonstrated the superiority by providing the model with examples containing the original problem decomposed into multiple sub-problems along with their answers~\citep{least-to-most}. However, \citet{juneja-etal-2023-small} illustrated that decoupling the decomposer from the solver by finetuning a separate decomposer language model (LM) to coordinate with a larger solver LM is beneficial to simply prompting a single monolithic LM to decompose and solve. Echoing their findings, \citet{wu2024divideorconquer} also found that distilling decomposition abilities from a larger LM to a smaller LM is much more generalizable compared to decomposing the solver abilities directly.
\begin{figure*}
    \centering
    \includegraphics[width=\textwidth]{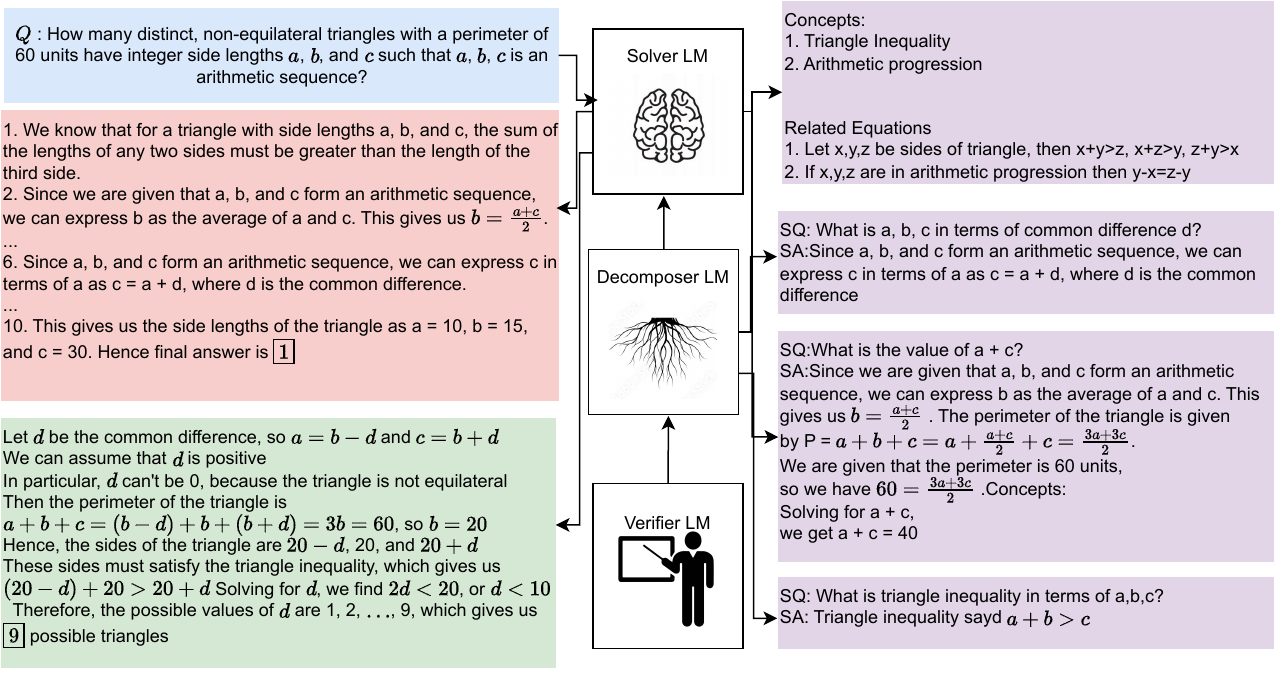}
    \caption{The inference procedure of \shortname{} on a question from the MATH dataset. A question (in \textcolor{CornflowerBlue}{blue}) is provided to the Solver LM that produces an incorrect answer (in \textcolor{RedOrange}{red}). The question is then provided to the Decomposer LM that generates the concepts and step-by-step subquestions (in \textcolor{Orchid}{lilac}). Each subquestion is answered by the Solver LM, and the sub-answer is verified by a Verifier LM. If the Verifier LM approves the sub-answer, that subqustion-subanswer pair is added to the context of reasoning steps; otherwise, a new subquestion is generated. The question, concepts, subquestions, and subanswers are provided in context to the Decomposer LM to generate the next subquestion. Finally, the question, concepts, subquestions, and subanswers are provided to the Solver LM to generate the final answer (in \textcolor{YellowGreen}{green}). }
    \label{fig:overall}
\end{figure*}

{\bf Our contributions.} However, a major bottleneck in existing methods of decomposer finetuning is the lack of tightness between the decomposer-solver interactions. Typically, the decomposition is done in a memoryless manner, with or without the solver's initial response; no strategy is employed to track whether the solver can follow the decomposed chain of reasoning. Towards this very end, we propose a novel multi-LLM coordination framework, {\bf L}anguage {\bf M}odel {\bf M}ultiplex (\shortname{}). \shortname{} is built upon three separate LMs, each dedicated to three different components of complex multistep reasoning -- a {\bf solver} LM is responsible for answering questions; a {\bf verifier} LM provides feedback on the correctness of the output from the solver, and a {\bf decomposer} LM identifies the basic concepts required to solve the problem and generates step-by-step subproblems by decomposing the original question (see Figure~\ref{fig:overall} for a working example). Unlike prior approaches, the decomposer in \shortname{} generates each subproblem depending on the solver's answers to prior subproblems, along with the verifier's feedback on those answers. Furthermore, the decomposer generates the conceptual requirements to solve the problem, which further streamlines the solver LM. Irrespective of the complexity of the underlying reasoning, the world knowledge required to answer any question is typically better preserved in larger, proprietary LMs. Considering this, we 
use GPT-3.5 ({\tt text-davinci-003}) as the solver without finetuning. For both the decomposer and verifier, we implement parameter-efficient fine-tuning~\citep{LoRA} of LLaMA-2 (13 billion parameters) separately. First, these models are finetuned separately towards the tasks of decomposition and verification using datasets annotated by GPT-4. The decomposer is then taught to coordinate with the solver and the verifier models in a policy learning setup. \shortname{} achieves promising performance across a diverse set of reasoning tasks. On the MATH dataset of mathematical reasoning, \shortname{} outperforms the best decomposer-tuning baseline by a staggering margin 8.1\% of absolute accuracy on average. Although \shortname{} uses the training split of the MATH dataset for tuning the decomposer and the solver, it seamlessly generalizes to out-of-distribution tasks in MedQA and JEEBench, outperforming the best competitive baseline with 9.7 \% and 7.71\% difference on absolute accuracy respectively. 

Beyond the discourse of overall numbers, we perform in-depth ablation analyses to identify the roles of each component of the model. We observe that (i) the verifier LM and concept generated by the decomposer LM play a crucial role in generalizing out-of-distribution reasoning tasks like MedQA, JEEBench Chemistry, etc.; (ii) finetuning the decomposer is crucial for better concept identification -- finetuned LLaMA-2 7B generates more effective conceptual requirements compared to even GPT-4; (iii)  even while not using all the modular components of \shortname{}, the prompt template of structured reasoning boosts the performance of GPT-4.

\section{Related Work}
\label{sec:related-work}

The efficacy of explicitly generating intermediate reasoning steps over direct generation of the required answer was first demonstrated by \citet{nye2021scratchpad}. Chain-of-thought prompting~\citep{CoT} generalized the scratchpad learning of \citet{nye2021scratchpad} into an in-context learning regime using LLMs. Chain-of-thought and its successors~\citep{chen2022program, yao2024tree} typically let the decomposition of a composite, multi-step reasoning problem remain implicit in the LLM.

\citet{least-to-most} demonstrated that instead, an explicit call to the LLM to generate multiple smaller problems that are steps to answer the original query achieves more robust reasoning. Their proposed method, Least-to-Most prompting, uses these simpler subproblems and their answers as the context to solve the original problem. Similarly, \citet{khot2023decomposed} proposed a prompting-based problem decomposition approach where the LLM is asked to decompose a complex task using few-shot examples. However, this still burdens a single language model in handling both decomposition and solution. \citet{juneja-etal-2023-small} circumvented this challenge by distilling the decomposition abilities into a relatively smaller language model. Their proposed method, DaSLaM, utilizes two separate language models that coordinate with each other to solve complex reasoning problems. Their findings suggest that finetuning the decomposer is more generalizable than finetuning the solver model. This has been further supported by \citet{wu2024divideorconquer} recently. \citet{tarasov2024distilling} explored the distillation of decomposition abilities via offline reinforcement learning. 
\citet{DSP} proposed a programmatic retrieval augmentation framework, namely Demonstrate-Search-Predict (DSP), for knowledge-intensive generation tasks. DSP relies on the coordination between a generative LM and a retrieval model through sophisticated programs. Recent attempts have been made to incorporate dense verifiers (typically, a finetuned, bidirectional language model acting as a classifier) aiding a generative model towards robust, verifiable problem solving and text generation~\citep{cobbe2021training, sun2023towards}. Different techniques for verification of LM-generated outputs have been proposed subsequently, such as self-verification~\citep{weng2023large}, majority voting~\citep{li2023making}, etc.

\section{Methodology}

Our proposed method, \shortname{}, is built upon the coordination of multiple LMs to perform reasoning in a modular fashion. However, such coordination is not implicit in the pertaining stage of a model; instead, we seek to inculcate this ability via finetuning (parts of) the LM multiplex. To this end, \shortname{} is built upon three functional components: a (preferably larger) solver model, a decomposer model, and a verifier model.

For fine-grained control over the function of the different components of \shortname{}, we make use of a structured, step-by-step input-output framework (see Figure~\ref{fig:overall}). The role of each of the modules in \shortname{} is described as follows.


\subsection{Decomposer}
\label{subsec:decomposer}

The \texttt{decomposer} LM guides the solver LM to solve a multi-step reasoning question in two ways. First, it provides the solver model with a set of concepts required to solve the problem. Second, it tells the solver LM what is the next sub-question required to solve given the previous sub-questions and their answers. More specifically, the decomposer LM is a function that can be defined as $D(q, \{s_i, sa_i\}, c): Q  \times  S  \times  SA \rightarrow \{S, C\}$, where $q$ represents the initial question to be solved,
$\{s_i, sa_i\}$ denotes the set of previous sub-questions ($s_i$) and their corresponding answers ($sa_i$), and
($c$) signifies whether the function needs to predict the concept or the next sub-question. $Q$ is the space of all the questions, $S$ is the space of all sub-questions, $SA$ is the space of all sub-answers, and $C$ is the space of all concepts.

{\bf Supervised finetuning.} The decomposer training is performed in two stages similar to \cite{juneja-etal-2023-small}. The first stage is supervised finetuning, where the language model is finetuned on a dataset prepared using GPT-4. To create the dataset, we provided GPT-4 with a question and its gold reasoning. It was then asked to first generate all the concepts required to solve the question, followed by sub-questions and sub-answers. Only the questions that were answered correctly were included in the dataset. Each sample in the dataset can be expressed as a tuple $\{Q, c, \{s_i, sa_i\}_{i=1}^n, s_{n+1}\}$, where $s_{n+1}$ is the next sub-question given the previous sub-questions and answers. The decomposer was then finetuned on the standard language modelling objective. 

{\bf Policy optimization.} With the supervised finetuning step, the decomposer LM is conditioned to respond to reasoning problems with concepts and decomposed subquestions. However, it is still not able to take the feedback from the solver and the verifier models into account. To this end, we utilize Proximal Policy Optimization~\citep{ppo} with the decomposer as the policy and the solver and the verifier model as a black-box environment. Precisely, we compute different types of rewards utilizing the feedback from the verifier model that takes the solver model's response into account at each step and provides the decomposer with necessary refinement signals. 

\subsection{Verifier}
\label{subsec:verifier}

Given the complexity of multistep reasoning, we need the verifier to be able to provide nuanced feedback to the decomposer on the possible mistakes made by the solver; a binary correct/incorrect message as employed by prior works with verifiers~\citep{li2023making, weng2023large} will limit the decomposer model's scope of vision. For fine-grained control, the verifier is finetuned on a supervised dataset containing a question, an answer with an error made in the correct answer, a classification for the type of error, and an explanation for the classification. The verifier classifies the given input into nine classes as follows: \circled{1} {Conceptual mistakes}, \circled{2} {Computational mistakes}, \circled{3} {Procedural mistakes}, \circled{4} { Misunderstood question}, \circled{5} {Mistake in the first step}, \circled{6} {Mistake in first half}, \circled{7} {Mistake in second half}, \circled{8} {Mistake in last step}, and \circled{9} {No mistake}. The dataset was produced using GPT-4, asking it to generate an explanation for the classification given the correct solution, wrong solution and the classification. The verifier is finetuned to generate the explanation and the classification (see Section~\ref{sec:RLFH} for examples of each type of error message and explanation).

\subsection{Training with Decomposer Feedback}
\label{sec:RLFH}
The training dataset curated for the decomposer LM consists of only the correct answers; hence, the decomposer is blind to the possible errors that the language model can make. In order to make the decomposer generate meaningful questions, we further finetune the decomposer while working in synergy with the solver language model using Policy gradient methods.

\textbf{Environment.}
The environment consists of a black-box solver model $\Theta$. The model $\Theta$ generates an answer to the current question given the concepts and previous questions and their answers.

\textbf{Policy, action and state space.}
The decomposer language model $\phi$ comprises the policy network. A state $s$ in the state space $S$ is defined by the concatenation of the initial state $s_0$ and all the actions taken from the initial state to the current state. The initial state $s_0$ is defined as the initial question $Q$. The action space is defined as the token space of the language model $\phi$. Hence, a state $s_n$ can be represented as $(s_0, \{a_i\}_{i=1}^n)$, where $a_i$ is the action taken at the $i_{th}$ time step.

\textbf{Reward function.}
The reward is based on the feedback given by the verifier at each sub-question produced by the decomposer. The reward structure is intuitively designed to impose penalties for errors occurring in earlier sub-questions relative to those occurring in later ones. This is because fixing an early mistake can significantly increase the chances of the question being correct. Further, the policy is penalised more for conceptual and procedural mistakes as compared to computational mistakes. We construct the reward function for the $k^{th}$ sub-question as follows: 
\begin{equation}
    R = \gamma^{k} \sum_{i=1}^{9}{r_i}
\end{equation}
where $\gamma<1$ is the discount factor responsible for imposing more penalties on the earlier generations. $r_i$ are the rewards for individual feedback given by the verifier as defined below (for each type of reward, we provide an example question asked by the decomposer, an erroneous answer to that question by the solver, type of error identified and the explanation generated by the verifier in red textboxes). 

{\em Conceptual correctness reward} is defined as, 
\begin{equation}
    r_1 = -0.15 I[V(s_k, sa_k)=1]
\end{equation}
where $I$ is the indicator function, $V$ is the verifier that takes in input the $k^{th}$ sub-question ($s_k$) and its answer produced by the solver ($sa_k$) and outputs the category of mistake. 
This reward accounts for any mistake made by the solver in understanding or while applying a concept incorrectly. 
\begin{tcolorbox}[colback = red!30!white, boxsep=2pt,left=2pt,right=2pt,top=0pt,bottom=0pt]
 {\fontsize{8pt}{9.6pt}\selectfont 
   \(\mathbf{Q}\): What is \(a\), \(b\), \(c\) in terms of common difference \(d\)?\\
   \(\mathbf{A: }\)Since \(a\), \(b\), and \(c\) form an arithmetic sequence, we can express \(c\) in terms of \(a\) as \(c = a + d\), where \(d\) is the common difference.\\
   $\mathbf{Verifier}$: Conceptual Mistakes: The student misunderstood the concept of finding the possible values of the common difference in an arithmetic sequence to form a triangle with a given perimeter.}
\end{tcolorbox}

{\em Computational correctness reward} is defined as, 
\begin{equation}
    r_2 = -0.05 I[V(s_k, sa_k)=2]
\end{equation}
This reward accounts for cases where the model has made any mistakes in calculation or formula application. These kinds of mistakes are weighted less compared to all other types because an external tool can be easily used for these mechanical tasks.    
\begin{tcolorbox}[colback = red!30!white, boxsep=2pt,left=2pt,right=2pt,top=0pt,bottom=0pt]
 {\fontsize{8pt}{9.6pt}\selectfont 
    $\mathbf{Q}:$ What is the value of $a+c$?\\
    $\mathbf{A}:$ Since we are given that $a$, $b$, and $c$ form an arithmetic sequence, we can express $b$ as the average of $a$ and $c$. This gives us $b = \frac{a + c}{2}$. The perimeter of the triangle is given by $P = a + b + c = a + \frac{a + c}{2} + c = \frac{3a + 3c}{2}$. We are given that the perimeter is 60 units, so we have $60 = \frac{3a + 3c}{2}$. Solving for $a + c$, we get $a + c = 20$\\
    $\mathbf{Verifier:}$ Calculation Mistakes: The model has made a calculation mistake while calculating the value of $a+c$. $a+c$ should be 40.}
\end{tcolorbox}

{\em Procedural correctness reward} is defined as, 
\begin{equation}
    r_3 = -0.15 I[V(s_k, sa_k)=3]
\end{equation}
This encompasses any mistake made by the model while following a step-by-step procedure whose information is already given in-context while solving the problem. 
\begin{tcolorbox}[colback = red!30!white, boxsep=2pt,left=2pt,right=2pt,top=0pt,bottom=0pt]
 {\fontsize{8pt}{9.6pt}\selectfont 
    $\mathbf{Q}:$ What are the possible values of $a,d$?\\
    $\mathbf{A}:$ Substituting this into the equation $a + c = 20$, we get $a + (a + d) = 20$, which simplifies to $2a + d = 20 - a$. Since the side lengths must be integers, $a$ and $d$ must be integers. We can then try different values of $a$ and $d$ that satisfy the equation $2a + d = 20 - a$. After trying different values, we find that the only solution that gives integer side lengths is $a = 10$ and $d = 20$.
    $\mathbf{Verifier:}$ Procedural Mistakes: The student made a mistake in the procedure of solving the problem by not accurately finding the possible values of the common difference.}
\end{tcolorbox}

\textit{Question understanding reward} is defined as, 
\begin{equation}
    r_4 = -0.2 I[V(s_k, sa_k)=4]
\end{equation}
This reward accounts for the cases when the model has made a mistake in understanding the sub-question. Either it has not answered the quantity asked in the sub-question. This gives a signal that the sub-question was either incoherent with the previous questions or was too complex for the model to answer. This kind of mistake is important to address and, hence, is given a higher weight.
\begin{tcolorbox}[colback = red!30!white, boxsep=2pt,left=2pt,right=2pt,top=0pt,bottom=0pt]
 {\fontsize{8pt}{9.6pt}\selectfont 
    $\mathbf{Q}:$ How many distinct values of $a,b,c$ are possible?\\
    $\mathbf{A}:$ This gives us the side lengths of the triangle as $a = 10$, $b = 15$, and $c = 30$.\\
    $\mathbf{Verifier:}$ Mistakes Understanding Question: The model has made a mistake by not giving the number of distinct values.}
\end{tcolorbox}

\textbf{Reward based on place of mistake.} As discussed above, later mistakes are penalised less than the earlier ones. Hence, if a mistake is made in the first step, it is given a reward of $-0.2$. If the model makes a mistake in the first half of the sub-answer, it is given a reward of $-0.12$. For a mistake in the last half of the sub-answer, it is given a reward of $-0.08$.  If the mistake is made in the last step, it is given a reward of $-0.05$.

\textbf{No-mistake rseward} is the case when the model has not made any mistake in answering the sub-question and is given a positive reward of $+1$.

\subsection{Inference}
During the inference, the decomposer, solver, and verifier models work together to answer a given question (see working example in Figure~\ref{fig:overall}). During the inference, the decomposer first produces a list of concepts required to solve the question. Then, given the question and concepts as context, the decomposer produces a sub-question. The sub-question is answered by the solver. Now, given the sub-question and sub-answer, the verifier provides feedback in the form of a multi-class classification into the above-described classes of mistakes. If the feedback provided by the verifier consists of either a conceptual mistake, procedural mistake, mistake in understanding or mistake in the first step, we again generate the sub-question.

\section{Experiments}
\label{sec:experiments}

For all the experiments, \shortname{} uses the OpenAI {\tt text-davinci-003} model (hereafter mentioned as GPT-3.5) as the solver and LLaMA-2 13B~\citep{llama} as the base models for the decomposer and the verifier.

\subsection{Training data curation}

For the first stage of finetuning of the decomposer LM, we curated a dataset of $15,396$ question, concept, sub-question, sub-answer tuples. The questions were taken from the train split of the MATH dataset~\citep{MATH}. The questions were taken from the MATH dataset. For verifier LM finetuning, a dataset of $3,674$ question-answer-classification tuples was generated. Details of the prompts used for each of these steps are provided in Appendix~\ref{app:data}.

\subsection{Training details}
We finetune LLaMA2-13B for both the decomposer and verifier. We train for 8 epochs with a batch size of 128, learning rate 2e-5, warmup steps of 100, a Lora r value of 4, LoRA Alpha of 16 and dropout of 0.05.  The models were trained in 8-bit quantization on an 80G A100 GPU.

For the second stage of fine-tuning, we finetuned the last 3 layers of LoRA adapters, using a batch size of 16, gradient accumulation steps=4, init kl coef=0.01, target=4. For inference, we used a temperature of 0 in all experiments for consistency of results with a max output length of 2000.

\subsection{Evaluation}
We evaluate our method on hard reasoning datasets that require multi-step reasoning. These datasets include MATH~\citep{MATH} (test split), JEEBench~\citep{jeebench}, and MedQA~\citep{medqa} (English questions). The MATH dataset contains math questions from challenging math competitions, since it was also used for training, this shows our performance on in-domain questions. Next, we evaluate on the out-of-distribution datasets like JEEBench which contains PCM questions extracted from the JEE Advanced exam and MedQA which contains open-domain questions from professional medical board exams. We only evaluate questions in the English language.

\subsection{Baseline Details}

We compare \shortname{} with five existing methods: Chain-of-thought prompting ({\bf CoT})~\citep{CoT}, Least-to-most prompting ({\bf L2M})~\citep{least-to-most}, Progressive Hint Prompting ({\bf PHP})~\citep{PHP}, Demonstrate-Search-Predict ({\bf DSP})~\citep{DSP}, and {\bf DaSLaM}~\citep{juneja-etal-2023-small}. The original setting of PHP requires an 8-shot prompting; however, since all other methods including \shortname{} predict in the zero-shot setting, we use PHP in 1-shot for a fairer comparison.

\subsection{Ablation Study}
\label{subsec:ablation}

In our investigation, we perform five types of ablation studies aimed at comprehensively understanding the significance of each component within the \textit{\shortname{}} pipeline.

We start with investigating the relevance of the verifier by conducting an experiment where we remove it entirely (\shortname{}\(\setminus\)V). Here, we accept each question generated by the decomposer during the inference process without any verification. 
Then, we explore the role of concepts within the pipeline. Here, we alter the approach by instructing the decomposer to directly generate sub-questions, without providing the concepts to the Solver LM during the answer generation phase (\shortname{}\(\setminus\)C). 
Following this, we investigate the incremental gains achieved through the second stage of finetuning via policy learning. To accomplish this, we analyze the performance of the decomposer checkpoint after the initial stage of fine-tuning, referred to as (\shortname{}\(\setminus\)RL).

To assess the impact of different types of rewards provided, we partition the rewards into two distinct categories: i) based on the type of mistake, which encompasses conceptual, computational, procedural, and question understanding correctness, and ii) based on the position of mistake. Subsequently, we come up with two ablation variants, finetuned using each category of rewards: \shortname{}-Type and \shortname{}-Position.
\begin{table}[!t]
\centering
\footnotesize
\setlength{\tabcolsep}{4pt}
\begin{tabular}{l|c|c|c|c|c|c}
\hline
    \multirow{2}{*}{Dataset} & \multicolumn{5}{c}{Method} \\\cline{2-7}
    & CoT & L2M & PHP & DSP & 
    DaSLaM & \shortname{}\\
    \hline
    PnC & \(16.4\) & \(16.0\) & \(10.2\) & \(16.2\) & 
    \(21.4\) &\(\bf 30.0\) \\
    NT & \(14.4\) & \(11.0\) & \(9.8\) & \(20.3\) & 
    \( 26.1 \) & \( \bf 41.0\) \\
    ALG & \(27.6\) & \(22.4\) & \(24.0\) & \(15.3\) 
    &\(33.4\) &\(\bf34.0\)\\
    I-ALG & \(16.4\) & \(16.8\) & \(10.0\) & \(17.0\) 
    &\(24.8\) &\(\bf27.8\) \\
    Calc. & \(14.0\) & \(14.58\) & \(14.28\) & \(18.8\) 
    &\(18.2\)&\(\bf34.0\)\\
    P-ALG & \(32.3\) & \(28.0\) & \(26.5\) & \(28.0\) 
    & \(44.0\)& \(\bf47.0\)\\
    Geom. & \(14.2\) & \(12.5\) & \(14.0\) & \(5.2\) 
    & \(21.4\)& \(\bf 32.0\)\\
    \hline
    MedQA & \(50.3\) & \(49.8\)& \(47.5\)&\(52.3\)
    &\(50.1\)&\(\bf 57.1\) \\
    \hline
\end{tabular}
\caption{Performance comparison of \shortname{} with the baselines on  MATH and MedQA datasets using GPT-3.5 as the solver LM.}
\label{tab:2}
\end{table}

\begin{table*}[!t]
\centering
\footnotesize
\setlength{\tabcolsep}{5pt}
\begin{tabular}{lccccccccccccc}
    \hline
    \multirow{2}{*}{Method} & \multicolumn{11}{c}{Dataset} \\\cline{2-13}
    & \makecell{Phy.\\ MCQ} & \makecell{Math.\\ MCQ} & \makecell{Phy.\\ Multi.} & \makecell{Math.\\ Multi.} & \makecell{Phy.\\ Num.} & \makecell{Math.\\ Num.} & \makecell{Phy.\\ Int.} & \makecell{Math.\\ Int.}& \makecell{Chem.\\ Int.}  & \makecell{Chem.\\ Num.} & \makecell{Chem.\\ Multi.} & \makecell{Chem.\\ MCQ}\\
     \hline
     CoT & 33.33 & 21.9 & 6.25 &12.0 & 3.03&1.69 &12.5 &10.8 &17.3&11.6&11.6&40.0\\ 
     PHP & 22.22& 17.07	& 6.25	&	7.59 &3.03	& 1.69	&	0*&	 4.0 &11.7 & 9.7 & 12.2 & 37.5 \\ 
     L2M & 22.22&	21.9&	6.25&	12.5&	3.03&	3.38&	10.0& 10.8 &13.0&9.7& 10.0&20.0	\\ 
     DaSLaM & \underline{55.5} & 29.5	& 18.7 & 16.0 & 	6.06 & 10.1 &  15.7 & 11.7& 14.2 & 9.2 & 11.6&14.6\\
      GPT4 & {\bf 55.5} & {\bf 34.1} &  {\bf 27.5} & {\bf 21.5} & \underline{15.1} &  \underline{11.8} &  {\bf 22.7} & {\bf 24.3} & \underline{17.9} & {\bf 25.5}& {\bf 48.3} & {\bf 60.0}\\
     \shortname{} & 51.85 & \underline{30.18} & \underline{26.8} & \underline{16.4} & {\bf 15.15}& {\bf 13.1} & \underline{16.2} & \underline{13.5} & {\bf 26.0} & \underline{23.2} & \underline{26.6} & \underline{53.3}\\ 
     \hline
     \shortname{}\(\setminus\)V & 37.03 & 24.52 & 14.6& 11.7& 12.2 &11.4&11.4&11.7&17.3&16.2&13.3&30.0\\
     \shortname{}\(\setminus\)C & 29.62& 20.75 &14.6&9.4&9.09&10.8&9.0&8.1&17.3 & 11.6& 13.3& 16.6\\
     GPT4-C&29.62&28.3&14.6&11.5&15.15&11.4&9.0& 11.4&21.7&23.2&33.33&30.0\\
     \shortname{}\(\setminus\)RL & 33.33 &21.9&18.7&12.7&12.2&10.1&10.0 &8.1&17.3&12.4&13.3 & 27.3\\
     \shortname{}-Type & 46.1 & 28.0 & 20.3& 14.0& 13.4& 11.4& 15.0& 13.5& 24.0& 23.2& 23.6& 45.4\\
     \shortname{}-Position & 38.4 & 24.52 & 16.0 & 12.9& 12.2& 11.4& 15.0& 10.8& 24.0& 20.6& 20.3& 33.0\\     
     \hline
     GPT35-SP  & 33.3 & 29.2	& 7.5 &  12.6& 	9.0 & 8.4 &  12.5 & 8.0& 17.6 & 9.2 & 12.2& 41.6 \\
      GPT4-SP & 61.1 & 36.5 & 30.0 & 26.5 & 30.0 &14.2 &43.75 & 32.0 & 17.6 & 36.5 & 49.1&66.6\\
      \hline\hline
\end{tabular}
    \caption{{\bf Performance of \shortname{} on JEEBench Dataset along with baselines and ablation variants}. ({\em Top third}) we highlight best and second best methods in {\bf boldface} and \underline{underline}. \shortname{} generally outperforms all existing prompting techniques with GPT-3.5 on different topics and different types of questions (other than Physics MCQ questions). In 3/12 cases, \shortname{} outperforms GPT-4. ({\em Middle third}) we observe a large drop in performance with each ablation variant, pointing towards an efficient integration of these modules into \shortname{} (see Section~\ref{subsec:ablation} for the description of each variant). ({\em Bottom third}) Performance of the structured answer generation employed in \shortname{}, without decomposer and verifier, using GPT-3.5 and GPT-4 as solvers.}
    \label{tab:jeebench}
\end{table*}
\section{Results}
\label{sec:results}
We summarize the performance of \shortname{} along with the baseline methods on the MATH and MedQA datasets in Table~\ref{tab:2} and on the JEEBench dataset in Table~\ref{tab:jeebench}. Across all the datasets, \shortname{} improves upon existing methods (using GPT-3.5 solver) by a huge margin. It demonstrates an average \(8\%\) improvement on the MATH dataset and an average \(2.5\%\) improvement on the JEEBench dataset as compared to the best-performing baseline DaSLaM.

\textbf{Can it improve on out-of-domain tasks?} In both DaSLaM and \shortname{}, the solver model is kept frozen with the hope of retaining generalizability. However, the decomposer model in both methods (and the verifier in \shortname{}) are finetuned using mathematical reasoning problems. This raises the question of the generalizability of these finetuned components over problems other than mathematical reasoning.
One of the most significant challenges with DaSLaM is that it is not able to perform well on out-of-domain tasks like JEEBench Chemistry. We find that our method can surpass this limitation as can be seen in Tables~\ref{tab:2} (MedQA) and \ref{tab:jeebench} (JEEBench Chemistry). While DaSLaM degrades the performance over CoT on MedQA, \shortname{} achieves an absolute accuracy gain of \(6.8\) percentage points. 

\textbf{How important is the verifier?}
Next, we seek to investigate the relative importance of each component in our pipeline. We observe that the accuracy decreases substantially upon removing the verifier model (\shortname{}\(\setminus\)V in the middle third of Table~\ref{tab:jeebench}). We can see that there is a drop of \(13.0\%\) in Chemistry versus \(10.08\%\) in Physics and \(3.4\%\) in Math subsets. The relative drop in accuracy with the ablation of the verifier is sharper with multi-answer, numeric, and integer answer questions. This makes sense given the computational reasoning requirement is higher in these problems and the verifier plays a crucial role in guiding the decomposer and the solver along the correct reasoning path. 
\begin{figure}[!h]
    \centering
    \includegraphics[width = 0.98\linewidth]{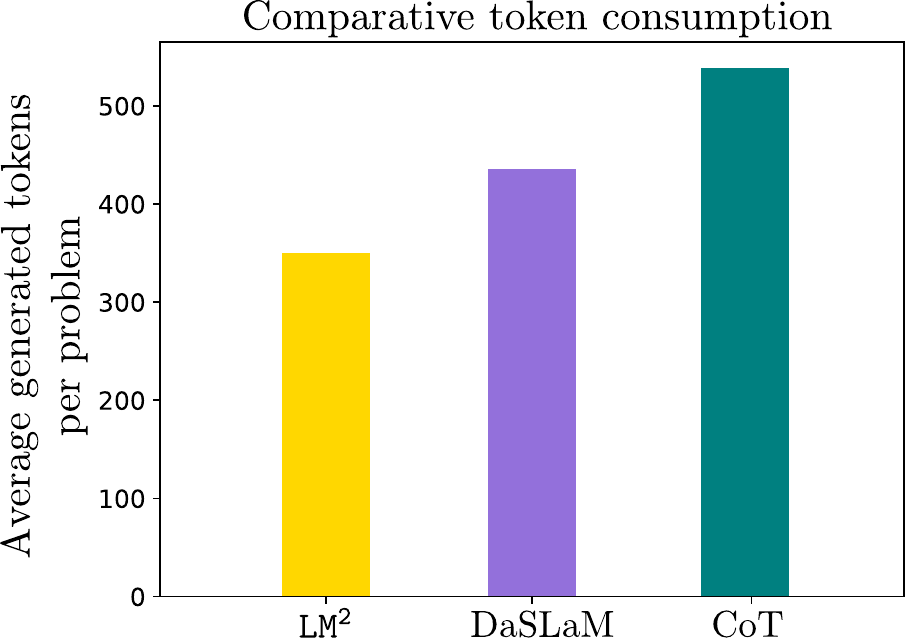}
    \caption{{\bf Comparison of token generation cost.} We depict the average number of tokens generated by the solver model using different methods to solve the given question averaged over 50 questions from the JEEBench dataset.}
    \label{fig:tokens}
\end{figure}
\begin{figure*}[!t]
    \centering
    \includegraphics[width = \textwidth]{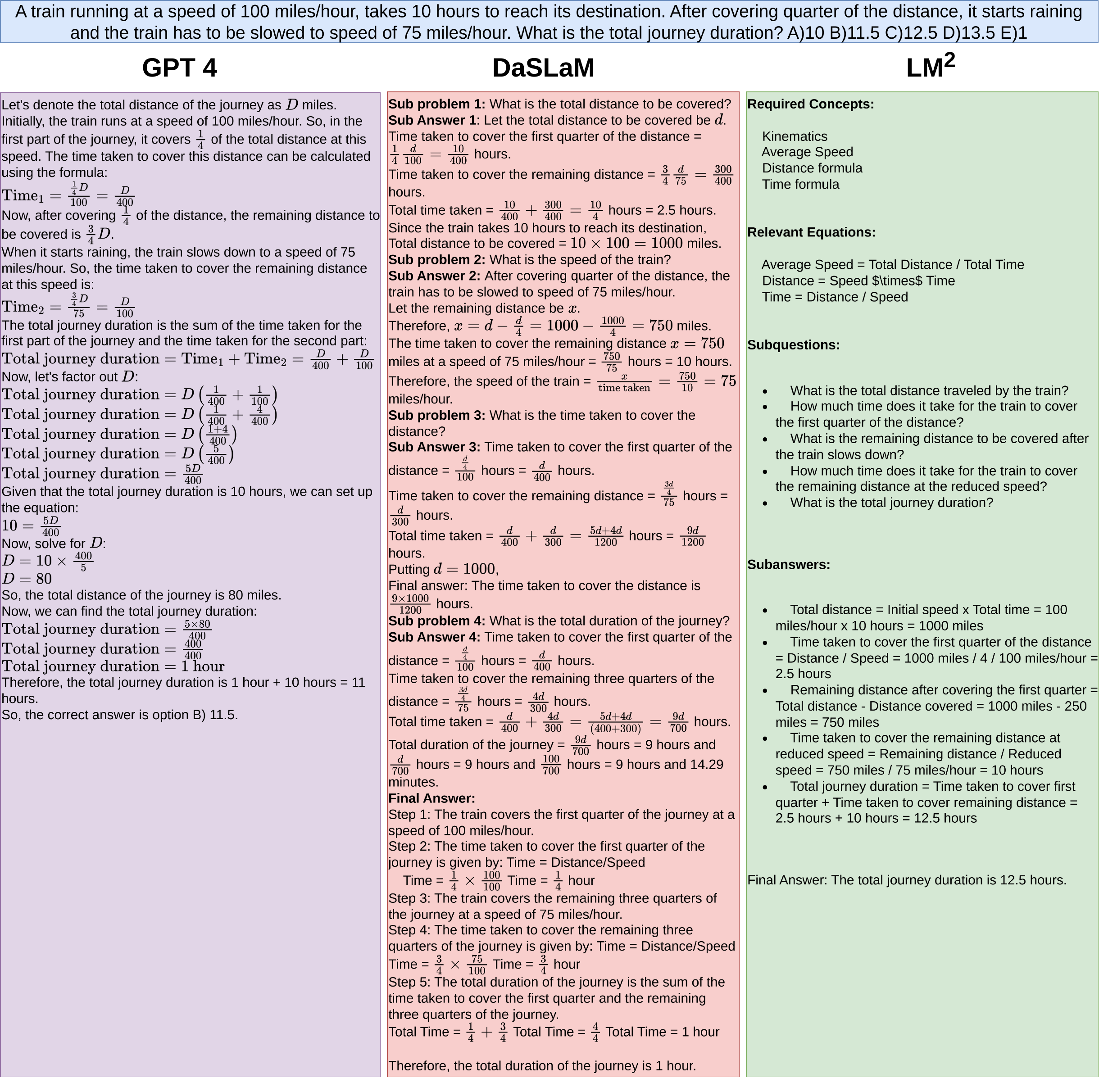}
    \caption{Comparison of GPT-4, DaSLaM and \shortname{} on an example from MATH dataset.}
    \label{fig:Example}
\end{figure*}

\textbf{How important are the concepts?} 
As can be seen from Table~\ref{tab:jeebench}, removing concepts decreases the accuracy of Physics subset by \(11.6\%\), Maths subset by \(6.03\%\), and Chemistry subset by \(17.5\%\). This shows that concepts also play a very important role in improving the performance on out-of-domain datasets like Physics and Chemistry. Typically, \shortname{}\(\setminus\)C fares worse than the rest of the ablation variants, demonstrating that the concepts are the most important component in \shortname{}.

{\bf GPT-4 as concept generator?} We also check how our decomposer compares to GPT-4 while generating concepts. To compare this, we prompt GPT-4 to generate concepts given the question. We observe that there is an average decrease of \(9.13\%\) when generating concepts using GPT-4 when compared to the Decomposer model, indicating the higher quality of concepts generated as a result of feedback-based fine-tuning.

\textbf{What is the effect of feedback-based finetuning?}
The effect of feedback-based fine-tuning is evident when comparing the performance of the decomposer without the second stage of fine-tuning alongside the verifier to that of \shortname{}. On average, we observe a notable decrease of \(9.6\%\) in performance when the second stage of fine-tuning is omitted. This finding highlights the significance of fine-tuning as a crucial step in optimizing model performance. However, the importance of concepts and the verifier appears to outweigh that of fine-tuning. This suggests that while fine-tuning contributes to improved model performance, the incorporation of concepts and a verifier into the model architecture yields more substantial enhancements.

\textbf{How does the structured answering template contribute?}
Recall that in \shortname{}, we introduce a novel, structured answering template for controllable coordination between the three models. It is imperative to investigate the role of such a template alone behind the performance boost. We make use of the template with two different solver models, GPT-3.5 and GPT-4.
As we can see in the bottom third of Table \ref{tab:jeebench} (coined as {\tt modelname}-SP), both models improve upon their base performance with our structured template. However, the stronger GPT-4 model is able to utilize the template much more efficiently, with an average gain of \(7.8\%\) across the JEEBench problems. Typically, improvement on Physics problems is higher than the Math problems, indicating that language models are not very good at retrieving physics concepts and solving the problem when using chain-of-thought prompting. It should noted that while the structured answering template alone is a powerful boost, it is much weaker alone without the complete coordination in \shortname{}.

{\bf Does guided reasoning help limit token usage?} An important challenge with iteratively interacting with an LLM is the increased token usage that will translate to expenses in either computational or monetary terms. In Figure~\ref{fig:tokens}, we plot the average token usage (per problem) incurred by the solver model (GPT-3.5) while using \shortname{} and DaSLaM against that of base chain-of-thought generation. Note that we only show the token usage corresponding to the modified responses while using \shortname{} and DaSLaM. Both these methods originally use base CoT to generate the initial response and therefore, their total token usage will always be higher than that of CoT. However, the added structure and guided reasoning significantly reduce the token usage in the modified response. \shortname{} prevails in this aspect too. A major reason behind this is the step-by-step synergy between the decomposer, the solver, and the verifier in \shortname{}. Since the decomposer generates the subquestion depending upon the response from the solver to the previous subquestion, the chances of redundant generation decrease, as opposed to DaSLaM where the subquestions are generated all at once.

{\bf Example analysis.} To further understand the nuances of \shortname{}, we perform an analysis of the generated output on an example from the MATH dataset (see Figure~\ref{fig:Example}).
We compare between \shortname{}, DaSLaM and GPT-4 with CoT. As we can see, GPT-4 makes an incorrect interpretation of the question itself. It assumes that the total journey after delay takes 10 hours, leading to an incorrect choice of option. 
The subquestions produced by DaSLaM do not adhere to the order of reasoning required to solve the problem and generate redundant questions. It starts with asking {\em What is the total distance to be covered?} However, in the second question, it asks for the speed of the train which is already given in the question itself. The 3rd subquestion generated by DaSLaM is actually the original question, and the solver makes a numerical mistake by simplifying the fraction \(\frac{\frac{3d}{4}}{75}\) to \(\frac{d}{300}\) instead of \(\frac{d}{100}\). Without a verifier, this erroneous response is integrated into the reasoning context of the solver. In the next questions, the same problem is asked to be solved and the solver continues to make incorrect responses. With \shortname{}, we observe a much more well-defined, crisp line of questioning by the decomposer model; the solver is able to reach the correct answer option without regenerating the same information or drawing incorrect subanswers.

\section{Conclusion}
\label{sec:conclude}

In this paper, we present \shortname{}, a cooperative cohort of generative language models working together to solve complex reasoning problems. \shortname{} utilizes a frozen solver model that is guided to solve reasoning problems by incrementally answering questions framed by a decomposer model and checked by the verifier model that is trained to coordinate with each other. We find that \shortname{} proves its supremacy over existing methods over a variety of reasoning tasks, both in-domain and out-domain. We find that despite being trained using mathematical reasoning examples, our proposed structured response scheme along with the fine-grained verification strategy plays a crucial role in generalizing \shortname{} to heavily out-of-distribution tasks like medical question answering and chemistry.

{\bf Limitations.} Despite promising results, \shortname{} bears some inherent limitations. Compared to purely prompting-based methods, it requires a certain computational overhead for the two-staged training. With proprietary LLM-based solvers, \shortname{} incurs extra token usage over single-pass solutions like chain-of-thought. Implicit limitations of the solver model, like lack of length generalization, arbitrary digit manipulation, etc. are expected to be inherited in \shortname{} as well. A possible future work can be towards incorporating deterministic solvers and tools into the multiplex.

\bibliography{custom}

\appendix

\section{Training Data Creation}
\label{app:data}
The data was generated using GPT-4. A temperature of 0.7 is used to ensure diversity in the generated data. We only stored the sub-question, sub-answer dataset if the number of sub-questions generated was more than three, this was done to ensure high data quality so that the model is able to decompose longer and more difficult questions effectively. First, we generate all the concepts, then the sub-questions given the question and the gold chain of thought. Finally, we generate the sub-answer given the question, a gold chain of thought and the sub-question to be answered. for the verifier, we first ask the LLM to answer the given question using standard COT prompting. Then based on the correctness of the answer, we take the solution chain of thought produced by the LLM and the gold answer and ask the LLM to classify the produced solution based on the mistake made. If the answer is correct, we store it separately and include it to make up to 10\% of the dataset with the label as 'No Mistake'. Prompts for the data curation are given below.  
\label{sec:appendix}
\subsection{Verifier Data Creation}

\subsubsection{Prompt}
\texttt{
You are a teacher, and you are grading a student's answer to a question. The student's answer is as follows:
    \{COT\_LLM\}
    The correct answer is as follows:
    \{COT\_gold\}
    Please provide feedback to the students on the mistakes they have made. You need to fill out a rubric and classify the mistakes into the following categories:}

\texttt{1. Conceptual Mistakes: The student has misunderstood the concept or has applied the wrong concept.\\
    2. Computational Mistakes: The student has made a mistake in the calculations.\\
    3. Procedural Mistakes: The student has made a mistake in the procedure of solving the problem.\\
    4. Mistake in understanding the question: The student has made a mistake in understanding the question.\\
    5. Mistake in the first step: The student has made a mistake in the first step of the solution.\\
    6. Mistake in the first half: The student has made a mistake in the first half of the solution.\\
    7. Mistake in the second half: The student has made a mistake in the second half of the solution.\\
    8. Mistake in the last step: The student has made a mistake in the last step of the solution.\\
    9. No mistake: The student has not made any mistake.}

\texttt{Please first provide feedback then fill the rubric and then finally tell your feedback to the student in between <feedback> and </feedback> tags as shown below:\\
    For example, if you want to tell the student that they have made a mistake in the first step and a conceptual mistake, then you need to write the following:\\
    <feedback> 1,4 </feedback>
    Do not write anything else in between <feedback> and </feedback> tags except the numbers.}

\texttt{Now, please provide feedback to the student on the mistakes they have made.}

\subsection{Decomposer Data Creation}
\subsubsection{Concepts data creation}
\texttt{   I have a question's solution, tell me all the specific concepts, theorems and formulas (separated by a comma,) used in it. An example is given below. \\
    Question: How many primes are in the row of Pascal's Triangle that starts with a $1$ followed by a $6$?\\
    Answer: If the row contains a $1$, then a $6$, then the binomial coefficients must be $\binom{6}{0}$ and $\binom{6}{1}$. All we need to check now are $\binom{6}{2}$ and $\binom{6}{3}$, since $\binom{6}{0} = \binom{6}{6}$, $\binom{6}{1} = \binom{6}{5}$ , and $\binom{6}{2} = \binom{6}{4}$. $\binom{6}{2} = \frac{6!}{4! \times 2!} = 15$, and $\binom{6}{3} = \frac{6!}{3! \times 3!} = 20$. None of those is prime, so there are $\boxed{0}$ prime numbers in the given row.\\    Concepts: Coefficients in Pascal's Triangle, Binomial Coefficients Formula, Prime Numbers\\
    Question: {question}\\
    Answer: {answer}\\
    Concepts:}

\subsubsection{Sub-question data creation}
    \texttt{I have a question, it's a solution and a sub-question. \\
    Your task is to break the question into sub-questions based on the steps in the answer.\\
    Keep the following tips in mind:\\
        1. Make sure not to break the question into trivial sub-questions, the sub-questions should be informative. \\
        2. The sub-questions should not require multiple steps to answer, something like 2-3 steps to solve is ideal. \\
        3. One way to break the question could be to identify what all quantities are required in the question by observing it's answer and then try to frame sub-questions based on the unknown entities.\\
        4. Make sure to put each question in the question tag like \$ question(What is the acceleration of the car as a function of time?)\$\\
    One example is given below. \\
  Question: How many primes are in the row of Pascal's Triangle that starts with a $1$ followed by a $6$?\\
    Answer: If the row contains a $1$, then a $6$, then the binomial coefficients must be $\binom{6}{0}$ and $\binom{6}{1}$. All we need to check now are $\binom{6}{2}$ and $\binom{6}{3}$, since $\binom{6}{0} = \binom{6}{6}$, $\binom{6}{1} = \binom{6}{5}$ , and $\binom{6}{2} = \binom{6}{4}$. $\binom{6}{2} = \frac{6!}{4! \times 2!} = 15$, and $\binom{6}{3} = \frac{6!}{3! \times 3!} = 20$. None of those is prime, so there are $\boxed{0}$ prime numbers in the given row.\\   
    Sub-questions: \\
    \$ question(How can the first two numbers be represented in form of binomial coefficients?)\$, \$ question(What are the values of all the coefficients in the row?)\$, \$ question(How many of the above numbers are prime?)\$\\
    Question: {question}\\
    Answer: {answer}\\
    Sub-questions:}
\subsection{Sub-answer data generation}
\texttt{I have a question, it's solution and a sub-question.\\
        I want you to answer the subquestion along with an explanation. \\
        Make sure to put the sub-answer in the answer tag like \$sub-answer(The acceleration of the car at time t = 2 seconds is speed / time = 2m/s/2s = $1m/s^2$)\$ \\
        Think step by step. \\
        Question: {question}\\
        Answer: {answer}\\
        Sub-question: {sub-question-array[i]}\\
        Sub-Answer: }

\end{document}